\documentclass{article}
\usepackage{spconf,amsmath,graphicx}
\usepackage{amsmath}
\usepackage{amssymb}
\usepackage{booktabs}
\usepackage{multirow}
\usepackage{array}
\usepackage{paralist}
\usepackage{extra_styles}
\usepackage{textpos}
\usepackage[pagebackref,breaklinks,colorlinks]{hyperref}

\usepackage[capitalize]{cleveref}
\crefname{section}{Sec.}{Secs.}
\Crefname{section}{Section}{Sections}
\Crefname{table}{Table}{Tables}
\crefname{table}{Tab.}{Tabs.}

\usepackage[moderate]{savetrees}

\ninept
\title{Self-Sufficient Framework for Continuous Sign Language Recognition}

\name{
    \begin{tabular}{c}
        Youngjoon Jang$^{1}$, Youngtaek Oh$^{1}$, Jae Won Cho$^{1}$, Myungchul Kim$^{1}$, \\
        Dong-Jin Kim$^{2}$, In So Kweon$^{1}$, Joon Son Chung$^{1}$ 
    \end{tabular}
    \thanks{\hspace{-4pt}$^\dag$Corresponding authors}
}
\address{$^1$Korea Advanced Institute of Science and Technology, Daejeon, Republic of Korea \\       
          $^2$Hanyang University, Seoul, Republic of Korea}

\begin{document}
%
\maketitle

\begin{abstract}
The goal of this work is to develop \emph{self-sufficient} framework for Continuous Sign Language Recognition (CSLR) that addresses key issues of sign language recognition.
These include the need for
complex multi-scale features such as hands, face, and mouth for understanding, and absence of frame-level annotations.
To this end, we propose (1) Divide and Focus Convolution (DFConv) which extracts both manual and non-manual features without the need for additional networks or annotations, and (2) Dense Pseudo-Label Refinement (DPLR) which propagates non-spiky frame-level pseudo-labels by combining the ground truth gloss sequence labels with the predicted sequence.
We demonstrate that our model achieves state-of-the-art performance among RGB-based methods on large-scale CSLR benchmarks, PHOENIX-2014 and PHOENIX-2014-T, while showing comparable results with better efficiency when compared to other approaches that use multi-modality or extra annotations.
\end{abstract}
\begin{textblock*}{.8\textwidth}[.5,0](0.5\textwidth, -.365\textwidth)
\centering
{\small Project page with demo: \url{https://mm.kaist.ac.kr/projects/ssslr}}
\end{textblock*}
\section{Introduction}
\label{sec:intro}

The Continuous Sign Language Recognition (CSLR) task aims to recognise a gloss\footnote{Glosses are the smallest units having independent meaning in sign language.} sequence in a sign language video~\cite{cheng2020fully,huang2018video,koller2019weakly}.
To capture the meaning of the sign expressions from a signer, recent works obtain manual and non-manual expressions by fusing RGB with other modalities such as depth~\cite{molchanov2016online}, infrared maps~\cite{liu2017continuous} and optical flow~\cite{cui2019deep}, 
or by explicitly extracting multi-cue features~\cite{huang2018video,camgoz2017subunets,kim2021dense,kim2021acp++} or human keypoints~\cite{zhou2020spatial} using off-the-shelf detectors.
However, using such extra components introduce bottlenecks in both training and inference processes.
In addition, most CSLR datasets only have sentence-level gloss labels without frame- or gloss- level labels~\cite{huang2018video,cihan2018neural,koller2015continuous}. 
To overcome insufficient annotations, the Connectionist Temporal Classification (CTC)~\cite{graves2006connectionist} loss has been traditionally opted to consider all possible underlying alignments between the input and target sequence.
However, using the CTC loss without true frame-level supervision produces temporally spiky attention which can make the model fail to localise important temporal segments~\cite{min2021visual}.

Accordingly, we develop \emph{self-sufficient} framework for CSLR, which provides meaningful gloss supervision while capturing helpful multi-cue information \emph{without additional modalities} or \emph{annotations.}
To this end, we propose two novel methods: Divide and Focus Convolution (DFConv) and  Dense Pseudo-Label Refinement (DPLR).
DFConv is a task-aware convolutional layer which extracts visual multi-cue features by dividing spatial regions to focus on partially specialised features.
Note that DFConv is designed to leverage prior knowledge about the structure of human bodies without any additional networks or modalities.
In addition, DPLR elaborately refines an initially predicted gloss sequence from the model by referring a ground-truth gloss, and propagates frame-level gloss supervision without additional networks, unlike~\cite{cui2019deep,pu2018dilated}.
We emphasise that DPLR is generally applicable to other CSLR architectures or frameworks~\cite{cheng2020fully,min2021visual} to bring performance gain by reducing missing glosses in predictions.

\begin{figure}[t]
\centering
    \centering
    \includegraphics[width=.85\linewidth]{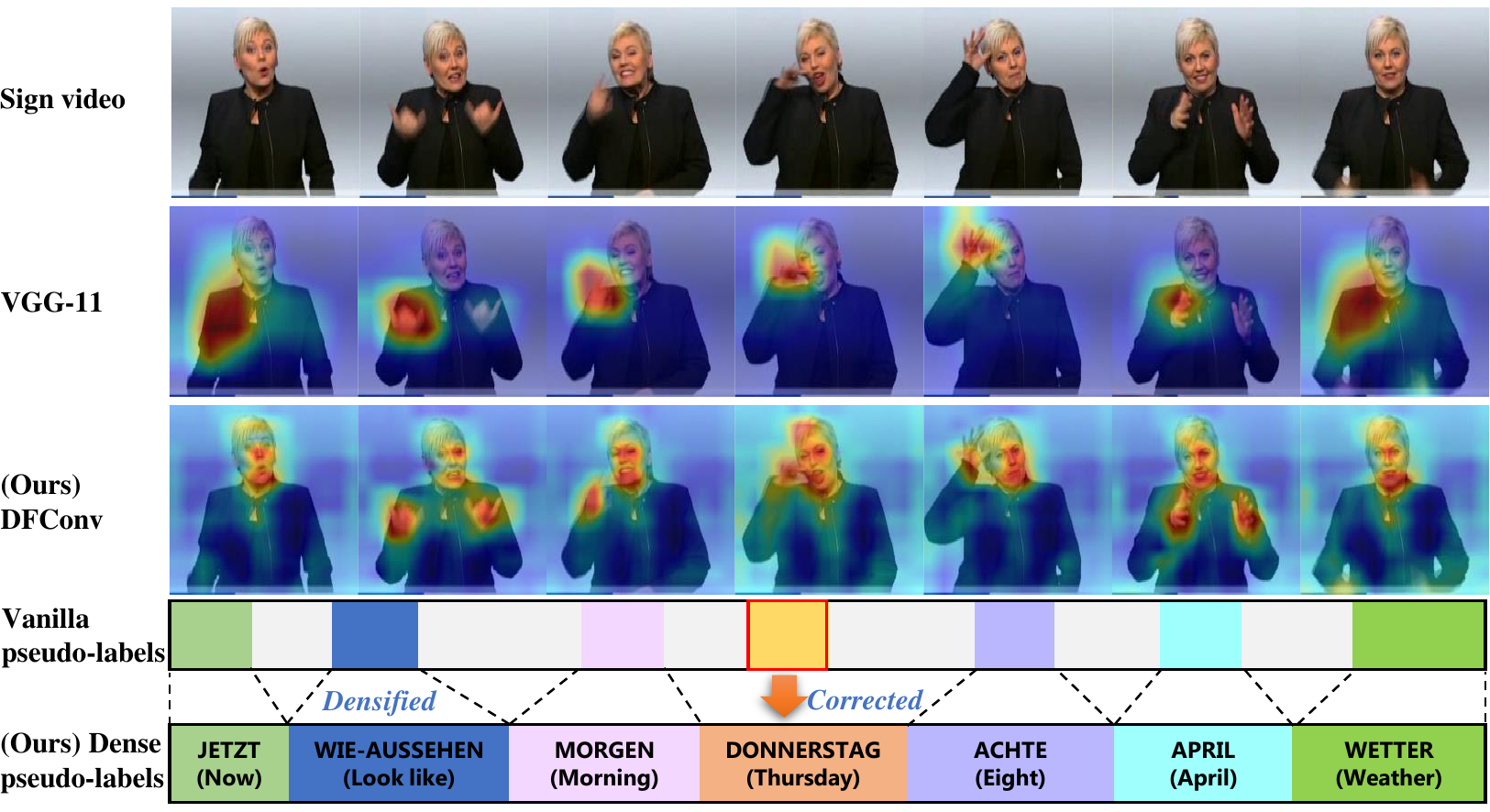}
    \vspace{-4mm}
    \caption{Comparison of GradCAM between VGG-11 and DFConv, and an example of the generated pseudo-labels before and after DPLR. DFConv better highlights multiple individual elements (hands, faces) across the entire scene whereas VGG-11 simply highlights a small region (\ie,~the right hand). DPLR corrects the mispredicted gloss with the ground truth gloss (\eg,~red box in Vanilla pseudo-labels) and densifies the pseudo-labels with the nearest glosses, which results in a more informative supervision without external knowledge.
    }
    \vspace{-5mm}
    \label{fig:teaser}
\end{figure}

We extensively validate the effectiveness of DFConv and DPLR. 
We also show that the whole \emph{self-sufficient} counterpart achieves state-of-the-art results among RGB-based methods and is comparable to other methods that use extra knowledge with better efficiency on two publicly available CSLR benchmarks~\cite{cihan2018neural,koller2015continuous}.
To summarise, our main contributions are as follows:

(1) We design a task-specific convolutional layer, named DFConv, that efficiently extracts non-manual and manual features without additional networks or annotations.
(2) We also introduce DPLR, a novel pseudo-label generation method, to propagate frame-level supervision by using the combination of the ground truth gloss sequence and the predicted temporal segmentation information.
(3) We conduct extensive experiments on two publicly available CSLR benchmarks, showing state-of-the-art performance compared to other RGB-based methods, and competitive results compared to other approaches that use multi-modality or additional knowledge {with better efficiency}.

\begin{figure*}[t]
\centering
    \includegraphics[width=.58\linewidth]{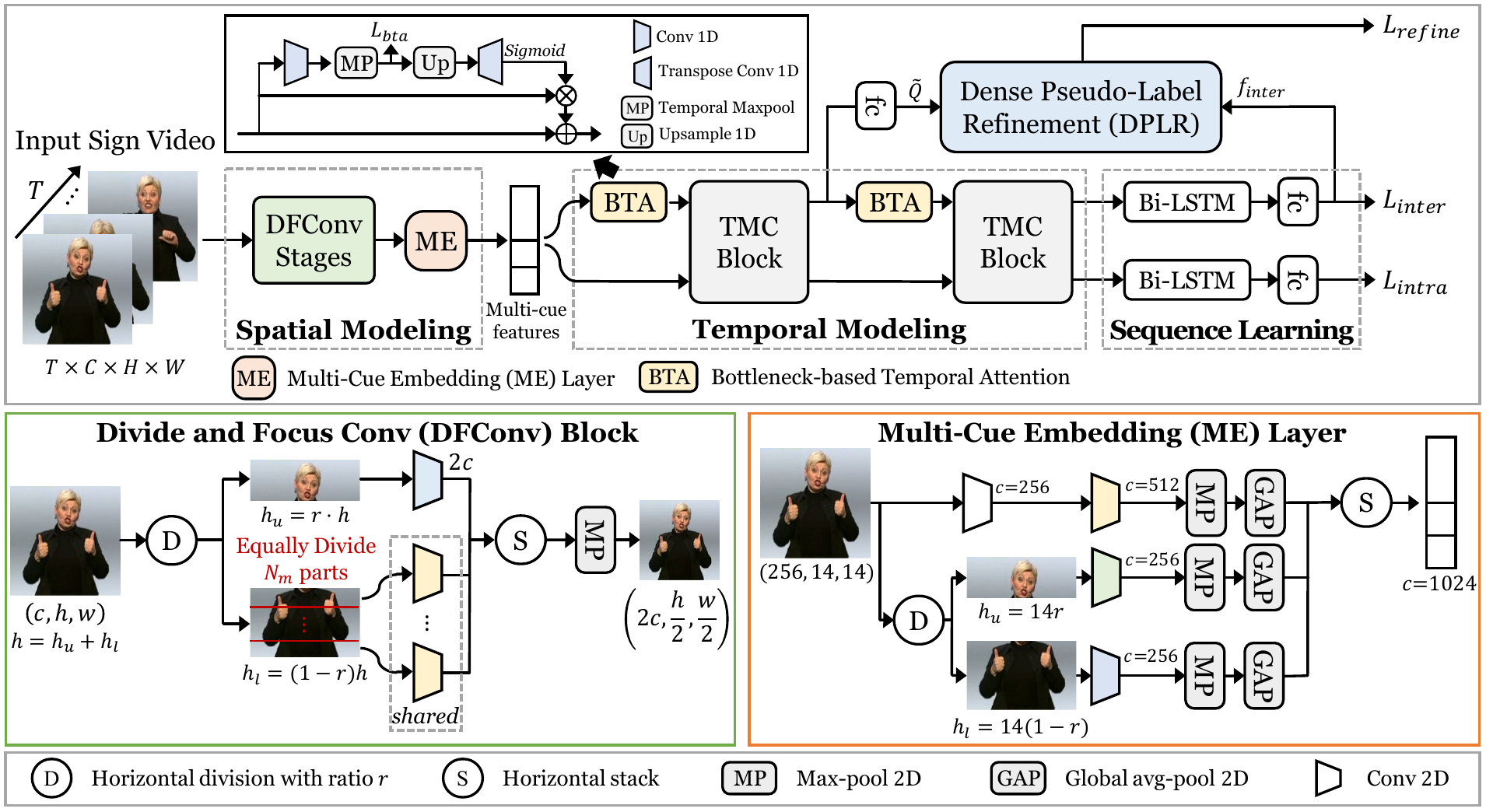}
    \vspace{-3mm}
    \caption{\textbf{Overall architecture}. 
    In Spatial Modeling, non-manual and manual features are extracted through DFConv followed by a Multi-Cue embedding layer. In Temporal Modeling, temporal features are extracted for gloss sequence prediction by integrating each element. Finally, the gloss probability vectors are obtained through sequence learning.}
    \label{fig:overall}
    \vspace{-5mm}
\end{figure*}

\section{Related works}
\label{sec:related}
Multi-cue fusion methods for CSLR task can be categorised into \emph{multi-semantic} and \emph{multi-modal} methods. 
Multi-semantic works~\cite{koller2019weakly,koller2015continuous,kim2021dense,kim2021acp++} utilised hand-crafted or weak-labeled features such as detected hands, trajectories of hands, and body parts, then integrate these features into frames to predict the gloss sequences. 
On the other hand, multi-modal works~\cite{liu2017continuous,molchanov2016online} use color, depth, and optical flow to extract orthogonal features. \cite{cui2019deep} proposed a multi-modality integration framework of appearance and motion cues by using both RGB frames and optical flow. Most recently, \cite{zhou2020spatial} fused human body keypoints extracted by an off-the-shelf network~\cite{sun2019deep}.
Unlike the methods listed above, we design DFConv that captures both manual and non-manual expressions from RGB video, without relying on \emph{any additional hand-crafted features} or \emph{multi-modal data}.

In addition, the CSLR task~\cite{huang2018video,camgoz2017subunets,pu2018dilated,koller2016deepA,jang2022signing} naturally corresponds to weakly-supervised learning problem due to
the lack of frame-level gloss annotations.
The challenge lies in the ambiguous semantic boundary of the adjacent glosses from sign videos~\cite{huang2018video,koller2015continuous,duarte2021how2sign}. To address this issue, 
some works in CSLR field generate frame-level pseudo-labels from sparse gloss annotations~\cite{koller2019weakly,koller2017re}, which can be inherently noisy and reliant on the model's performance. Most recently, the CTC loss~\cite{graves2006connectionist} is employed to facilitate end-to-end training of a deep learning model~\cite{cui2017recurrent,niu2020stochastic}, and consider all the feasible underlying alignments between the predictions and labels. However, as observed in~\cite{min2021visual,pu2019iterative}, directly optimising CTC can cause spiky attention in predictions, favoring more blank glosses. Recent works tackle this issue by balancing the blank output and meaningful glosses~\cite{cheng2020fully}, and by directly supervising the visual features via visual alignment constraint~\cite{min2021visual} and mutual knowledge transfer~\cite{hao2021self}.
In contrast, we propose Dense Pseudo-Label Refinement (DPLR) that provides dense and reliable supervision signals obtained by gloss predictions of the model to visual features. 

\section{Method}
\label{sec:overview}
CSLR task aims to map a given input video to its corresponding gloss sequence ${g}$ = $\{g_n\}{_{n=1}^N}$ with $N$ glosses.
As shown in \Fref{fig:overall}, 
a sign video is fed into the \emph{spatial modeling} module consisting of several Divide and Focus Convolution (DFConv) layers, and a multi-cue embedding layer to extract manual and non-manual features.
The multi-cue features of all the frames are 
passed through the \emph{temporal modeling} module, that is comprised of Bottleneck-based Temporal Attention (BTA), which captures more important information among adjacent frames,
and Temporal Multi-Cue (TMC) blocks of \cite{zhou2020spatial}.
Then, the output of last TMC block is passed through the \emph{sequence learning} stage, which is composed of a Bi-LSTM layer~\cite{schuster1997bidirectional} and FC layer to predict the gloss sequence from the final model output. 
Finally, the Dense Pseudo-Label Refinement (DPLR) module is introduced to effectively train the latent representations by generating corrected and densified frame-level pseudo-labels.

\subsection{Divide and Focus Convolution}
\label{sec:dfconv}
We observe from 
various
CSLR datasets~\cite{cihan2018neural,koller2015continuous,ham2021ksl} that non-manual expressions occur frequently in the upper region of the image, while manual expressions occur mainly in the lower region.
As shown in~\Fref{fig:teaser}, despite the importance of both non-manual and manual expressions appearing in the entire image area, the conventional 2D convolution layer tends to capture the only one most dominant information (\ie,~right hand) over the whole image.
To address this issue, we propose a novel Divide and Focus Convolution (DFConv) layer designed to independently capture non-manual features and manual features solely from RGB modality. 

The structure of the DFConv is illustrated in~\Fref{fig:overall}.
Inspired by the observation in~\cite{fan2020gaitpart}, DFConv physically limits the receptive field that increases as the network deepens by subdividing an image~\cite{kim2021dense} into upper (for non-manual expressions) and lower regions (for manual expressions) with the division ratio of $r$, where $r$ is the ratio of spatial height of the upper region $h_u$ to the original spatial height $h$ given by $r = \frac{h_u}{h}$.
To precisely capture the dynamic manual expressions, we further subdivide the lower region into $N_m$ groups.
For the upper region, this kind of subdivision is not required since non-manual expressions do not consist of the same amount of dynamics.
We empirically observe that subdividing the upper region reduces the performance as well.
Note that \emph{different convolution weights are used for each upper and lower regions}, and \emph{the weights are shared within the subdivided lower regions}.
This helps the model to focus more on visually meaningful areas that represent complex sign expressions in the segmented image.

Unlike other methods that leverage external knowledge, we only introduce two hyper-parameters $r$ and $N_m$, which make our method significantly more efficient.
By virtue of simply splitting the frames horizontally, DFConv efficiently captures multi-cue features simultaneously without equipping costly human pose estimator like STMC~\cite{zhou2020spatial} that increases model complexity and inference time (See project page).
To further embed the outputs of stage 3 in~\Fref{fig:overall} into the three individual multi-cue vectors (\ie, full-frame, non-manual and manual), a simple and effective Multi-Cue Embedding (ME) layer is employed.
The full-frame features containing global information are passed through two 2D convolution layers, and the remaining features (non-manual and manual) are passed through only single 2D convolution layer. Finally, all these features are vectorised by max pooling with a $2\times 2$ kernel followed by an average pooling layer.

\subsection{Dense Pseudo-Label Refinement}
\label{sec:dense_pl}
Most existing sign language datasets do not have temporally localised gloss labels~\cite{huang2018video,cihan2018neural,koller2015continuous,guo2018hierarchical}.
Due to the characteristics of the CTC loss used in training CSLR models without frame-level labels, the output sequence predictions of models are naturally induced to be sparse.
As a result, it is difficult for CSLR models to receive direct and precise alignment supervision for each gloss token.
In addition, without alignment supervision, CSLR models learn entire sequences as a whole instead of individual gloss words. 
This limits the robustness of models severely as they rely on entire sequences. 
In other words, models can easily confuse similar sequences with slightly different words.
In order to mitigate these drawbacks, we introduce an additional training objective called Dense Pseudo-Label Refinement (DPLR) that uses the alignment information predicted by the model to generate Dense Pseudo-Labels (DPL). Then, the model is further refined with these generated pseudo-labels.

\begin{figure}[!t]
    \centering
    \includegraphics[width=.70\linewidth]{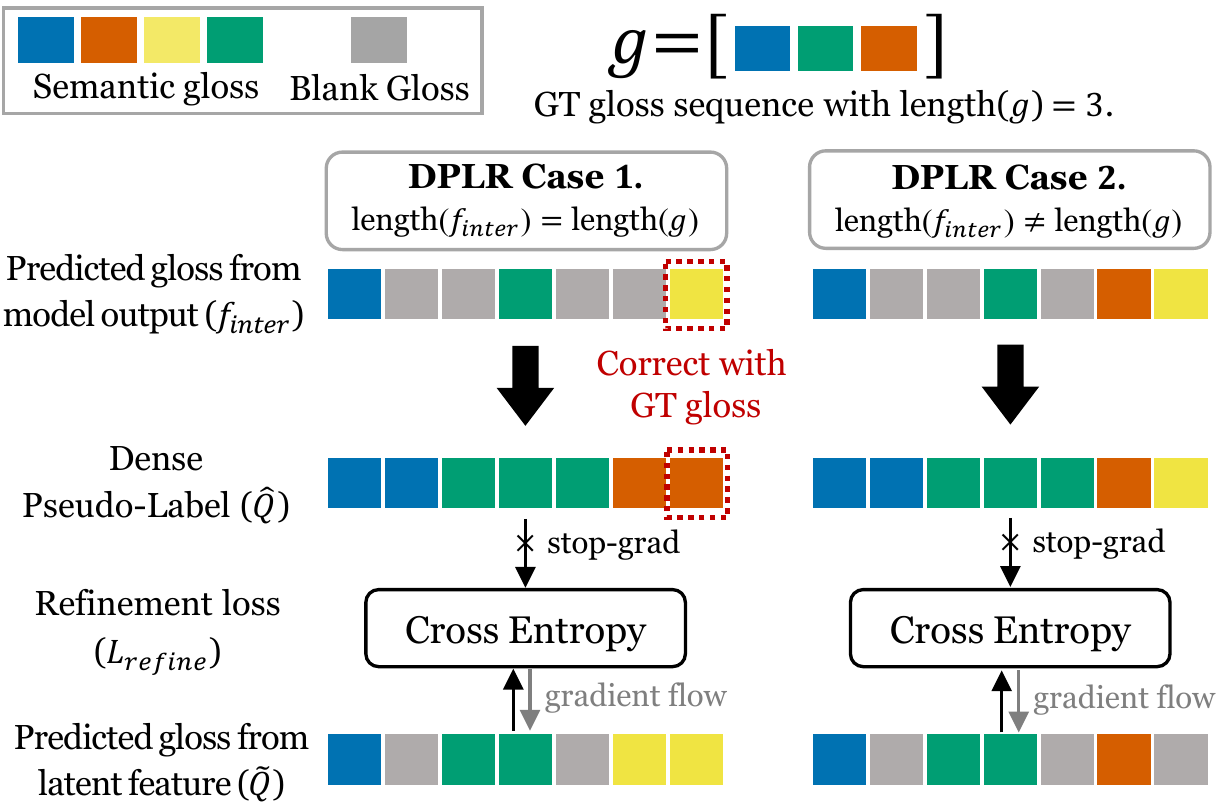}
    \vspace{-3mm}
    \caption{Dense pseudo-label generation process in DPLR. {\bf (Case 1)}: gloss sequence length predicted by the model and the ground truth are matched. {\bf (Case 2)}: gloss sequence length predicted by the model and the ground truth is off by one word.
    DPLR provides latent features with \emph{frame-level} supervision to compensate for the CTC loss while reducing the noise of pseudo-labels by the correction mechanism of Case 1.
    }
    \vspace{-5mm}
    \label{fig:DPLR}
\end{figure}

In DPLR process, we have two separate cases for generating DPL $\hat{Q}$ as illustrated in~\Fref{fig:DPLR}.
We first compare the sequence length of non-blank predictions of the model with its corresponding ground truth gloss sequence.
If the sequence length is matched, we go to {\bf Case 1}, where we compare the predicted gloss sequence with the ground truth sequence. If a predicted gloss is wrong, we swap in correct gloss from the ground truth to increase the reliability of the pseudo-labels. As mentioned before, the predictions are sparse due to the nature of the CTC loss, and most of the predictions along the temporal axis are blanks.
Here, we create DPL by filling each blank with the nearest predicted glosses.
In the case where the predicted sequence length differs from the ground truth by one gloss length, we go to {\bf Case 2}.
Then, we simply densify the pseudo-label using the nearest gloss without swapping any glosses regardless of the correctness of the glosses.
In the case that the sequence length differs by more than one gloss, we disregard that sequence as this might cause predictions of the model to degrade, so we do not propagate refinement loss $L_{refine}$.

Using pseudo-labels only from {\bf Case 1} and {\bf Case 2},
we refine the model with Cross Entropy (CE) loss on the latent features similar to~\cite{cheng2020fully} as follows:
\begin{equation} \label{eq:refine_loss}
    L_{refine} = \text{CE}(\hat{Q}, \tilde{Q}),
\end{equation}
where $\hat{Q}$ is dense pseudo-labels and, $\tilde{Q}$ is gloss probability acquired from latent features, which is the final output of the inter-cue path (See~\Fref{fig:overall}). 
Note that we demonstrate the efficacy of `Densify' and `Refine' processes in~\Cref{tab:abl_GRM}, and show that DPLR is generalisable to other models in 
our project page.

In addition, the quality of pseudo-labels generated from the model depend heavily on the model's performance. As the CSLR task aims to translate a sign language video into a gloss sequence by mapping several adjacent frames into one gloss, it is important to extract key frames in the video. 
Hence, we design the \textbf{Bottleneck-based Temporal Attention (BTA)} module to attend to the temporally salient frames among adjacent frames.
BTA consists of a temporal-wise attention map using 1D convolution layers and a max pooling layer to capture the temporally salient frames. The CTC loss is then propagated to the bottleneck, after the max pooled features, hence the name is Bottleneck.

With our additional modules, our final loss function is as follows:
\begin{align} \label{eq:total_loss}
    L_{total} = L_{inter} + \lambda_{1} L_{intra} + \lambda_{2} L_{refine} + \lambda_{3} L_{bta},
\end{align}
where, $L_{inter}$, $L_{intra}$ and $L_{bta}$ are all CTC losses. $L_{bta}$ is the average of all the TMC block's CTC losses and $\lambda_{1}$, $\lambda_{2}$, and $\lambda_{3}$ are loss weights.

\section{Experiments}
\label{sec:exp_results}
\noindent \textbf{Dataset and Evaluation Metric.}
We conduct experiments on two publicly available CSLR benchmarks to validate our \emph{self-sufficient} framework: PHOENIX-2014~\cite{koller2015continuous} and PHOENIX-2014-T~\cite{cihan2018neural}.
We adopt the Word Error Rate (WER)\footnote{WER = ({\#substitutions} + \text{\#deletions} + \text{\#insertions}) / ({\text{\#words in reference}})}~\cite{koller2015continuous} for evaluation. Furthermore, in our project page, we upload a demo video to visually demonstrate the effectiveness of DFConv.
\begin{table}[ht]
\centering
\begin{minipage}{0.47\columnwidth}
    \hspace*{\fill}
    \resizebox{.98\columnwidth}{!}{%
        \begin{tabular}{lccc}
            \toprule
             & Extra & \multicolumn{2}{c}{WER (\%)} \\
            Method & Annotations & Dev & Test \\
            \midrule
             DeepHand~\cite{koller2016deepA} & Hand & 47.1 & 45.1 \\
             SubUNets~\cite{camgoz2017subunets} & Hand & 40.8 & 40.7 \\
             Deep Sign~\cite{koller2016deepB} & Hand & 33.3 & 38.8 \\
             Staged-Opt~\cite{cui2017recurrent} & Hand & 39.4 & 38.7 \\
             LS-HAN~\cite{huang2018video} & Hand & - & 38.3 \\
             Align-iOpt~\cite{pu2019iterative} & - & 37.1 & 36.7 \\
             SF-Net~\cite{yang2019sf} & - & 35.6 & 34.9 \\
             DPD+TEM~\cite{zhou2019dynamic} & - &  35.6 & 34.5 \\
             cnn-lstm-hmm~\cite{koller2019weakly} & - & 27.5 & 28.3 \\
             Re-sign~\cite{koller2017re} & - & 27.1 & 26.8 \\
             DNF~\cite{cui2019deep} & - & 23.8 & 24.4 \\
             \bottomrule
        \end{tabular}%
    }
\end{minipage}
~
\begin{minipage}{0.48\columnwidth}
    \resizebox{.98\columnwidth}{!}{%
        \begin{tabular}{lccc}
            \toprule
             & Extra & \multicolumn{2}{c}{WER (\%)} \\
            Method & Annotations & Dev & Test \\
            \midrule
             cnn-lstm-hmm~\cite{koller2019weakly} & Mouth & 26.0 & 26.0 \\
             STMC~\cite{zhou2020spatial} & - & 25.0 & - \\
             SFL~\cite{niu2020stochastic} & - & 24.9 & 24.3 \\
             FCN~\cite{cheng2020fully} & - & 23.7 & 23.9 \\
             DNF+SBD-RL~\cite{wei2020semantic}\; & - & 23.4 & 23.5 \\
             DNF~\cite{cui2019deep} & Flow & 23.1 & 22.9 \\
             VAC~\cite{min2021visual} & - & 21.2 & 22.3 \\
             CMA~\cite{pu2020boosting} & - & 21.3 & 21.9 \\
             SMKD~\cite{hao2021self} & - & \textbf{20.8} & 21.0 \\
             STMC~\cite{zhou2020spatial} & Pose & {21.1} & \textbf{20.7} \\
             \midrule
             Ours & - & \textbf{20.9} & \textbf{20.8} \\
             \bottomrule
        \end{tabular}%
    }
\end{minipage}
\vspace{-3mm}
\caption{Comparison of performance in WER (\%) on PHOENIX-2014 benchmark. Ours shows the comparable performances to the existing state-of-the-art methods using either pose~\cite{zhou2020spatial} or algorithmic gloss segmentation~\cite{hao2021self} even without extra annotations.}
\label{tab:phoenix}
\vspace{-3mm}
\end{table}

\begin{table}[ht]
\centering
\begin{minipage}{0.47\linewidth}
    \hspace*{\fill}
    \resizebox{.98\linewidth}{!}{%
        \begin{tabular}{lccc}
            \toprule
            & Extra & \multicolumn{2}{c}{WER (\%)} \\
            Method &  Annotations  & Dev & Test \\
            \midrule
            cnn-lstm-hmm~\cite{koller2019weakly} & - & 24.5 & 26.5 \\
            FCN~\cite{cheng2020fully} & - & 23.3 & 25.1 \\
            SLRT~\cite{camgoz2020sign} & - & 24.9 & 24.6 \\
            SLRT~\cite{camgoz2020sign} & Text & 24.6 & 24.5 \\
            \bottomrule
        \end{tabular}%
    }
\end{minipage}
~
\begin{minipage}{0.48\linewidth}
    \resizebox{.98\linewidth}{!}{%
        \begin{tabular}{lccc}
            \toprule
             & Extra & \multicolumn{2}{c}{WER (\%)} \\
             Method & Annotations & Dev & Test \\
            \midrule
             cnn-lstm-hmm~\cite{koller2019weakly}\;\; & Mouth+Hand & 22.1 & 24.1 \\
             SMKD~\cite{hao2021self} & - & 20.8 & 22.4 \\
             STMC~\cite{zhou2020spatial} & Pose & \textbf{19.6} & \textbf{21.0} \\
             \midrule
             \textbf{Ours} & - & \textbf{20.5} & \textbf{22.3} \\
             \bottomrule
        \end{tabular}%
    }
\end{minipage}
\vspace{-3mm}
\caption{Comparison of performance in WER (\%) on PHOENIX-2014-T benchmark. Our framework achieves the state-of-the-art performances among RGB-based approaches, while shows comparable performances with the pose-based multi-cue method~\cite{zhou2020spatial}.}
\label{tab:phoenix_t}
\vspace{-6mm}
\end{table}

\begin{figure*}[t]
\vspace{-6mm}
   \centering
    \includegraphics[width=0.70\linewidth]{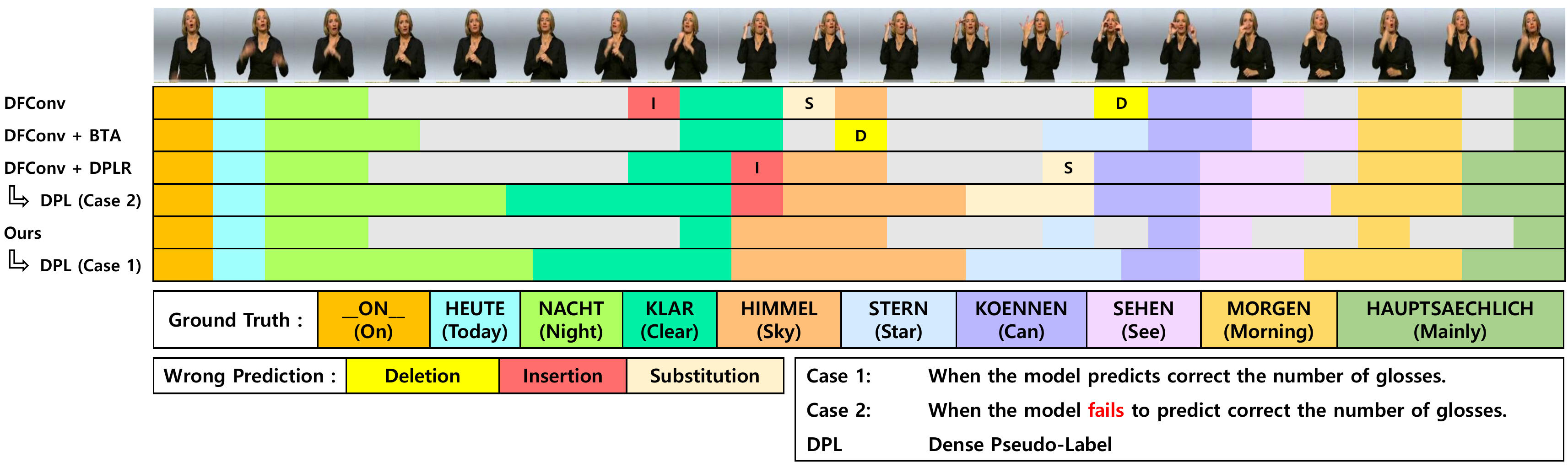}
   \vspace{-3mm}
   \caption{Gloss predictions in a single sentence sign video from different network architectures (D: deletion, I: insertion, S: substitution). In the fourth and sixth rows, we further visualise two cases of Dense Pseudo-Labels (DPL). Applying the DPLR on the prediction greatly reduces the deletion phenomenon. 
   }
   \label{fig:qualitative}
   \vspace{-6mm}
\end{figure*}

\subsection{Experimental Results}
\label{sec:quant_results}
We compare our framework with recent CSLR methods on both PHOENIX-2014~\cite{koller2015continuous} and PHOENIX-2014-T~\cite{cihan2018neural} benchmarks. \Cref{tab:phoenix,tab:phoenix_t} show the WER scores, 
while we specify the type of either extra annotations or modalities used during training for each method. 

\noindent \textbf{PHOENIX-2014.} 
\Tref{tab:phoenix} summarises the results on Dev and Test splits from PHOENIX-2014 for several CSLR baselines. 
First, Ours achieves the state-of-the-art performances on Test split among RGB-based approaches.
In particular, Ours outperforms the recently proposed FCN~\cite{cheng2020fully}, fine-grained labeling~\cite{niu2020stochastic}, VAC~\cite{min2021visual} with alignment supervision to visual features, and CMA~\cite{pu2020boosting} with both gloss and video augmentation.
Moreover, Ours shows superior performance over several recent methods that explicitly require extra annotations for training~\cite{huang2018video,koller2019weakly,cui2019deep}, and comparable performances to SMKD~\cite{hao2021self} with algorithmic gloss segmentations and STMC~\cite{zhou2020spatial} using pose annotations.
Note that the proposed method does not require either extra annotations for acquiring the benefit to detect spatially important regions or additional networks for the refinement of pseudo-labels.

\noindent \textbf{PHOENIX-2014-T.} \Tref{tab:phoenix_t} shows the results on Dev and Test splits of PHOENIX-2014-T. 
Ours surpasses cnn-lstm-hmm~\cite{koller2019weakly} which is trained with both mouth and hand annotations, and even outperforms SLRT~\cite{camgoz2020sign} that jointly learns sign recognition and translation task from both sign glosses and sentences.
Ours also outperforms SMKD~\cite{hao2021self}, a competing baseline using RGB modality, and shows comparable results to STMC~\cite{zhou2020spatial}.

\begin{table}[ht]
\centering
\begin{minipage}{0.475\linewidth}
    \centering
    \resizebox{.95\linewidth}{!}{%
        \begin{tabular}{ccccc}
            \toprule
            &  &  & \multicolumn{2}{c}{WER (\%)} \\
            \cmidrule{4-5}
            \;DFConv\; & \;\;DPLR\;\; & \;BTA\; & Dev & Test \\
            \midrule
             &  &  & 26.1 & 26.7 \\
             \cmidrule{1-5}
            \checkmark &  &  & 24.5 & 24.5 \\
            \checkmark & \checkmark &  & 22.4 & 22.5 \\
            \checkmark &  & \checkmark & 24.2 & 24.1 \\
            \cmidrule{1-5}
            \checkmark & \checkmark & \checkmark & \textbf{20.9} & \textbf{20.8} \\
            \bottomrule
        \end{tabular}%
    }
    \vspace{-2mm}
    \caption{Ablation study of DFConv, DPLR, and BTA. All the proposed components of our method gradually improve the performance. 
    }
    \label{tab:component}
\end{minipage}
\quad
\begin{minipage}{0.475\linewidth}
    \centering
    \resizebox{.95\linewidth}{!}{%
        \begin{tabular}{ccccc}
            \toprule
             &  &  &  \multicolumn{2}{c}{WER (\%)} \\
            \cmidrule{4-5}
            w/ $L_{refine}$ & Densify & Refine & Dev & Test \\
            \midrule
             &  &  & 24.2 & 24.1 \\
            \cmidrule{1-5}
            \checkmark &  &  & 23.5  & 23.8          \\
            \checkmark &  & \checkmark & 23.3 & 23.8 \\
            \checkmark & \checkmark &  & 22.4 & 22.5 \\ 
            \cmidrule{1-5}
            \checkmark & \checkmark & \checkmark & \textbf{20.9} & \textbf{20.8} \\
            \bottomrule
        \end{tabular}%
    } 
    \vspace{-2mm}
    \caption{Ablation study on the design choice of DPLR.
    Both `Densify' and `Refine' processes are key in improving performance.
    }
    \label{tab:abl_GRM}
\end{minipage}
\vspace{-4mm}
\end{table}

\subsection{Ablation Study}
\label{sec:abl_study}

\noindent \textbf{Component Analysis.}
In~\Tref{tab:component}, we ablate each component of our method to investigate its effectiveness. 
In the first row of the table, we show the result of the baseline model with VGG-11~\cite{simonyan2014very} architecture followed by three 1D convolution layers.
All components of our method consistently improve the performance altogether.
In particular, when BTA is combined with DFConv, the performance improvement is marginal, but when combined with DPLR, it shows a large performance improvement. From this, we conclude that DPLR and BTA are complementary modules to each other.
We ablate qualitatively the gloss predictions of each component in ~\Fref{fig:qualitative}. 

\noindent \textbf{Design Choice of DPLR.} 
The baseline in the first row of~\Tref{tab:abl_GRM} is the same baseline in the fourth row in~\Tref{tab:component}, which is the model trained with DFConv and BTA.
`Densify' and `Refine' 
indicate
whether the prediction from the model is filled by the nearest gloss prediction and whether glosses are replaced with ground truth glosses, respectively, as shown in~\Fref{fig:DPLR}.
We show from the second and third row (without `Densify') that directly leveraging the output of the model brings marginal improvements to the baseline.
`Densify,'
which provides direct alignment supervision on the \emph{frame-level} to the latent features is the key component for improving the model performance.
Finally, the proposed Dense Pseudo-Labels (DPL), which is the combination of both `Densify' and `Refine' processes, shows the best performance by the correction mechanism with ground truth labels in `Refine' to reduce the noise in \emph{dense} pseudo-labels.

\begin{figure}[ht]
  \centering
   \includegraphics[width=0.83\linewidth]{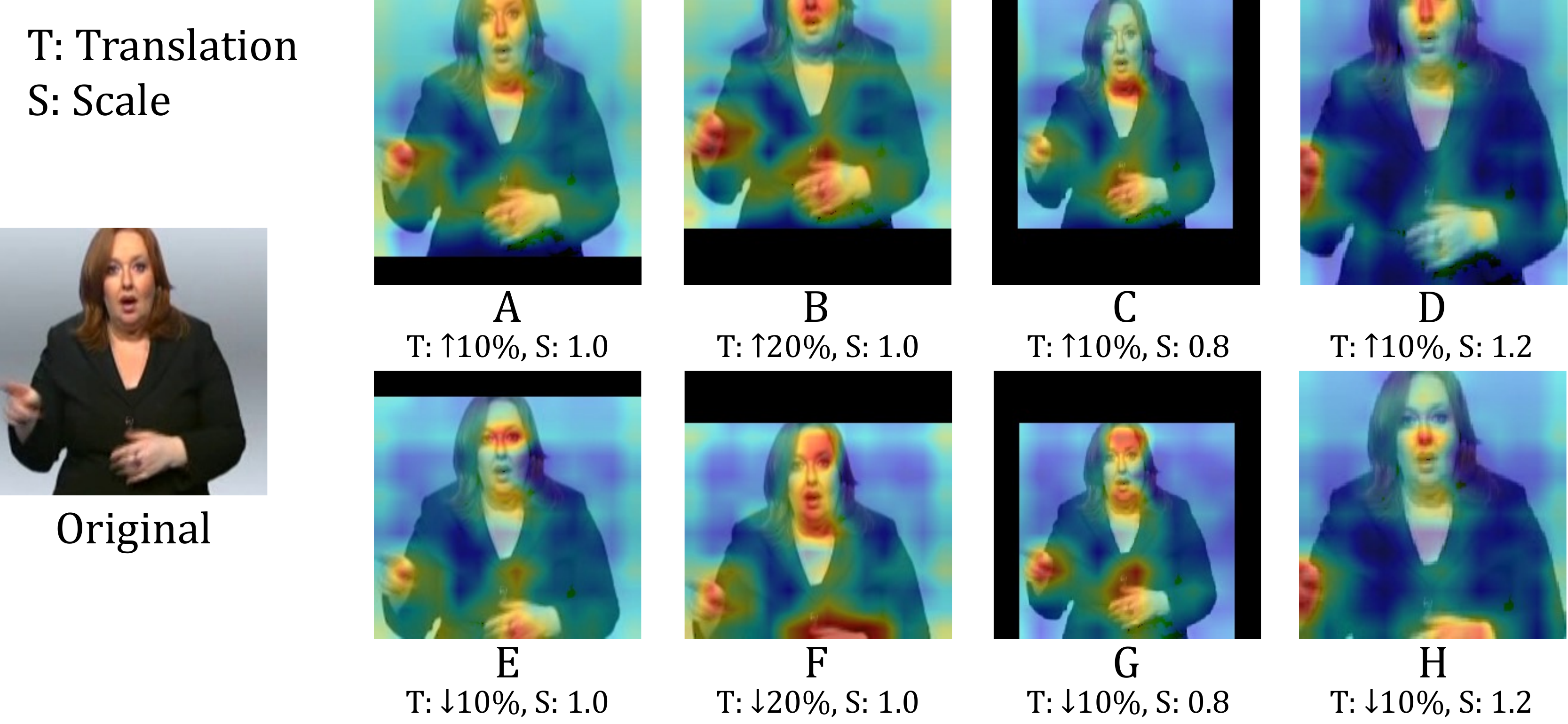}
   \vspace{-4mm}
   \caption{Activation maps from Ours on \emph{test-time} novel transformations.
   }
   \vspace{-2mm}
   \label{fig:transform}
\end{figure}

\begin{table}[ht]
    \centering
    \resizebox{0.75\linewidth}{!}{
        \begin{tabular}{ccccccccc}
        \toprule
         & \multicolumn{2}{c}{VAC} & \multicolumn{2}{c}{STMC} & \multicolumn{2}{c}{Ours} & \multicolumn{2}{c}{Ours$^\dagger$} \\
         \cmidrule(lr){2-3} \cmidrule(lr){4-5} \cmidrule(lr){6-7} \cmidrule(lr){8-9}
        Transform(T,S) & \multicolumn{1}{c}{Dev} & \multicolumn{1}{c}{Test} & \multicolumn{1}{c}{Dev} & \multicolumn{1}{c}{Test} & \multicolumn{1}{c}{Dev} & \multicolumn{1}{c}{Test} & \multicolumn{1}{c}{Dev} & \multicolumn{1}{c}{Test} \\
        \midrule
        Original                      & 21.2 & 22.3 & 21.0 & \textbf{20.7} & 20.9 & 20.8 & \textbf{20.9} & 20.8 \\
        A: ($\uparrow$10\%, 1.0)    & 23.9 & 24.1 & 22.9 & 21.3 & 22.9 & 23.0 & \textbf{21.1} & \textbf{21.2} \\
        B: ($\uparrow$20\%, 1.0)    & 27.1 & 27.1 & 33.6 & 32.9 & 24.2 & 24.9 & \textbf{22.7} & \textbf{22.5} \\
        C: ($\uparrow$10\%, 0.8)    & 38.5 & 38.0 & 42.9 & 41.1 & 32.0 & 30.9 & \textbf{28.5} & \textbf{28.1} \\
        D: ($\uparrow$10\%, 1.2)    & 28.9 & 29.8 & 31.4 & 32.0 & 26.9 & 27.1 & \textbf{24.8} & \textbf{24.5} \\
        \midrule
        E: ($\downarrow$10\%, 1.0)  & 30.9 & 30.4 & \textbf{24.4} & \textbf{24.1} & 27.2 & 26.8 & 24.9 & 24.6 \\
        F: ($\downarrow$20\%, 1.0)  & 35.5 & 34.7 & 31.4 & 30.0 & 31.1 & 30.9 & \textbf{29.7} & \textbf{29.9} \\
        G: ($\downarrow$10\%, 0.8)  & 56.5 & 52.3 & 46.7 & 42.1  & 46.5 & 43.9 & \textbf{38.1} & \textbf{37.9} \\
        H: ($\downarrow$10\%, 1.2)  & 28.5 & 27.5 & 26.9 & 28.0 & 26.2 & \textbf{26.2} & \textbf{26.1} & 26.3 \\
        \midrule
        Average & 33.7 & 33.0 & 32.5 & 31.4 & 29.6& 29.2 & \textbf{27.0} & \textbf{26.9} \\
        \bottomrule
        \end{tabular}
    }
    \vspace{-3mm}
    \caption{Robustness comparison with state-of-the-art methods in simulated real world scenario. We compare the WER on a model that has been trained on a train set without these transformations. Ours denotes a model with $r$ set to $0.35$, and Ours$^\dagger$ denotes a model where $r$ is moved along with the transformations during inference (T: vertical translation, S: scale).
    We note that STMC is our reproduction. We reimplement it as faithfully as possible.
    }
    \vspace{-3mm}
    \label{tab:division_ratio}
\end{table}
\noindent{\textbf{Robustness of DFConv.}}
\emph{Our method is more robust where the signer is not bounded to a specific region at inference time than the state-of-the-art methods in practical cases.}
To simulate such a scenario, we make a set of transformed data from PHOENIX-2014 Dev and Test splits, each of which includes a different degree of vertical translation (T) and scale operation (S).
In \Cref{tab:division_ratio}, we list RGB based state-of-the-art method (VAC), the pose-based method (STMC), our model tested with the original division ratio $(r=0.35)$ (Ours), and our model where the $r$ is changed along the corresponding transformation (Ours$^\dagger$).
Although shifting the $r$ gives the best performance (Ours$^\dagger$), in the real world, where we are not able to adjust on the fly $r$ (Ours), the average performance of Ours still surpasses VAC and STMC. 

We also present the activation maps of DFConv with the static division ratio $(r=0.35)$ in~\Cref{fig:transform}.
DFConv steadily captures non-manual and manual features even when half of the signer's face is out of focus (B, D) or when the signer's hands are partially out (D, F, H) with failure cases of pose-detectors shown in our project page.
This shows that pose-based sign recognition methods are heavily reliant on the performance of the pose-detector.

\section{Conclusion}
In this paper, we propose two novel methods, DFConv and DPLR, that complement missing annotations in the existing weakly-labeled sign language datasets. 
To the best of our knowledge, we are the first to propose a method to extract manual and non-manual features individually by designing a task-specific convolution without any additional networks or annotations.
In addition, we introduce DPLR module that does not require additional networks during the pseudo-labeling process and demonstrate its effectiveness through various experiments.
The experimental results show that our framework achieves state-of-the-art performance on two large-scale benchmarks among RGB-based methods, and also outperforms or is comparable to methods based on multi-modality.

\section{Acknowledgement}
This work was supported by Institute of Information \& communications Technology Planning \& Evaluation (IITP) grant funded by the Korea government (MSIT, 2022-0-00989, Development of Artificial Intelligence Technology for Multi-speaker Dialog Modeling for KAIST and 2020-0-01373, Artificial Intelligence Graduate School Program for Hanyang Univ.).


\bibliographystyle{IEEEbib}
\bibliography{shortstrings,strings}

\end{document}